\documentclass[10pt,twocolumn,letterpaper]{article}

\usepackage{times}
\usepackage{arxiv}

\usepackage[dvipsnames]{xcolor}
\usepackage{colortbl}
\usepackage{xspace}

\definecolor{dark_blue}{rgb}{0, 0, 0.7}

\definecolor{dark_green}{rgb}{0, 0.5, 0}

\usepackage[dvipsnames]{xcolor}
\usepackage{colortbl}
\usepackage{xspace}

\definecolor{dark_blue}{rgb}{0, 0, 0.7}

\definecolor{dark_green}{rgb}{0, 0.5, 0}

\definecolor{customblue}{rgb}{0.21,0.49,0.74}
\definecolor{changecolor1}{RGB}{255,0,0}

\usepackage{multirow}
\usepackage{colortbl}

\usepackage[breaklinks,colorlinks,
  linkcolor=customblue,citecolor=customblue,urlcolor=customblue]{hyperref}
\usepackage{graphicx}
\usepackage{capt-of}   
\usepackage{cuted}    
\usepackage{tikz}
\newcommand*\circled[1]{%
  \tikz[baseline=(char.base)]{
    \node[shape=circle, draw, inner sep=0.4pt, line width=0.3pt, font=\scriptsize] (char) {#1};
  }%
}

\usepackage{tabularx}
\usepackage{amsmath}
\usepackage{amssymb}
\usepackage{mathtools}
\usepackage{amsthm}
\usepackage{caption}
\usepackage{colortbl}
\usepackage{comment}
\usepackage{paralist}
\usepackage{tcolorbox}
\usepackage{ragged2e}
\usepackage{placeins}

\title{CorGi: Contribution-Guided Block-Wise Interval Caching\\ for Training-Free Acceleration of Diffusion Transformers}

\author{
Yonglak Son\textsuperscript{1}\thanks{Equal contribution}\quad
Suhyeok Kim\textsuperscript{1}\footnotemark[1]\quad
Seungryong Kim\textsuperscript{2}\thanks{Corresponding authors}\quad
Young Geun Kim\textsuperscript{1}\footnotemark[2]\\[10pt]
\textsuperscript{1}Korea University\quad
\textsuperscript{2}KAIST AI\\
{\tt\small\href{https://casl-ku.github.io/CorGi}{https://casl-ku.github.io/CorGi}}
}

\begin{document}
\maketitle

\begin{strip}
\centering
\includegraphics[
  width=\textwidth,
  height=\textheight,  
  keepaspectratio
]{Figure/Figure_teaser.pdf}

\vspace{-1pt}
\captionof{figure}{
\textbf{Teaser.} Our framework accelerates diffusion transformers (DiT) by 2.02\(\times\) on (a) Stable Diffusion 3.5-Large and by 1.91\(\times\) on (b) FLUX.1-dev, while preserving high image quality and consistency with images generated by the original model.}
\label{fig:teaser}
\end{strip}

\begin{abstract}

\vspace{-2pt}
Diffusion transformer (DiT) achieves remarkable performance in visual generation, but its iterative denoising process combined with larger capacity leads to a high inference cost. Recent works have demonstrated that the iterative denoising process of DiT models involves substantial redundant computation across steps. To effectively reduce the redundant computation in DiT, we propose CorGi (\textbf{Co}nt\textbf{r}ibution-\textbf{G}u\textbf{i}ded Block-Wise Interval Caching), training-free DiT inference acceleration framework that selectively reuses the outputs of transformer blocks in DiT across denoising steps. CorGi caches low-contribution blocks and reuses them in later steps within each interval to reduce redundant computation while preserving generation quality. For text-to-image tasks, we further propose CorGi+, which leverages per-block cross-attention maps to identify salient tokens and applies partial attention updates to protect important object details. Evaluation on the state-of-the-art DiT models demonstrates that CorGi and CorGi+ achieve up to 2.0$\times$ speedup on average, while preserving high generation quality.
\end{abstract}

\section{Introduction}

Diffusion models have achieved remarkable performance in visual domains, powering a wide range of vision tasks such as text-to-image generation~\cite{SD3.5, FLUX, chen2024pixart}, image-to-image translation~\cite{saharia2022palette, liu2025iidm}, and text-to-3D generation~\cite{poole2022dreamfusion, ding2024text}. 

Early diffusion models were built on U-Net backbones, as their encoder–decoder architecture with skip connections is effective for multi-scale denoising~\cite{rombach2022high, song2020denoising, ho2020denoising}. Recently, diffusion models are increasingly shifting to Diffusion Transformer (DiT) for their improved scaling behavior and higher representational capacity~\cite{esser2024scaling, zhou2024transfusion}. This transition has enabled the development of larger diffusion models, yielding higher generation quality and text-image alignment~\cite{peebles2023scalable, chen2024pixart}. In addition, multi-modal DiT (MM-DiT) further strengthens prompt adherence by processing joint self-attention over concatenated image-text tokens, contributing to state-of-the-art models~\cite{SD3.5, FLUX}.

Despite the advances in generation quality, DiT remains computationally inefficient. Its iterative denoising process, combined with a deep stack of transformer blocks and quadratic attention over long-range image–text tokens, leads to a high inference cost. To enable efficient inference, caching methods that reuse intermediate features have been widely explored. Existing caching methods can be broadly categorized into two approaches:  (1) block-segment caching, which reuses a subset of blocks from the previous step~\cite{chen2406delta,zhang2025blockdance,ma2024deepcache}, and (2) token-wise caching by selectively allocating computations for salient tokens~\cite{liu2502region,zou2024toca,cheng2025cat}. Although these methods can improve inference efficiency, they often skip salient information beyond redundancy, leading to quality degradation. 

To better understand the source of redundant computation, we first analyze the role and contributions of individual transformer blocks in DiT across denoising steps. We observe two key empirical findings: (1) blocks are spatially specialized and temporally varying in their contributions, and (2) at each step, only a small subset of blocks is highly active. To enable fine-grained removal of redundancy in denoising process based on our findings, in this paper, we propose \textbf{CorGi} {(\textbf{Co}nt\textbf{r}ibution-\textbf{G}u\textbf{i}ded Block-Wise Interval Caching)}, a training-free DiT inference acceleration framework which caches low-contribution blocks and reuses them in later steps within each interval that is composed of multiple steps.

CorGi begins with a short warm-up phase to establish the global structure, then performs block-wise interval caching. At each interval boundary, it calculates per-block contribution score via Centered Kernel Alignment (CKA)~\cite{kornblith2019similarity} between each block’s outputs at the initial step of consecutive intervals. Using contribution scores, CorGi caches low-contribution blocks and reuses their cached outputs in subsequent steps. For text-to-image (T2I) tasks, we further introduce CorGi+, which leverages per-block cross-attention maps to identify block-specific salient tokens and ensures partial attention updates on these tokens even when the block is cached to protect object details.

We evaluate CorGi and CorGi+ on state‑of‑the‑art DiT models --- Stable Diffusion 3.5 (SD3.5)~\cite{SD3.5}, FLUX.1~\cite{FLUX}, and PixArt-$\Sigma$~\cite{chen2024pixart} --- using the MS COCO dataset~\cite{lin2014microsoft}. Our evaluation results demonstrate that CorGi and CorGi+ achieve 2.0× end‑to‑end speedups, on average, achieving higher visual and semantic consistency with the original model outputs compared to the baselines.

\begin{figure}[t!]
\includegraphics[width=\linewidth]{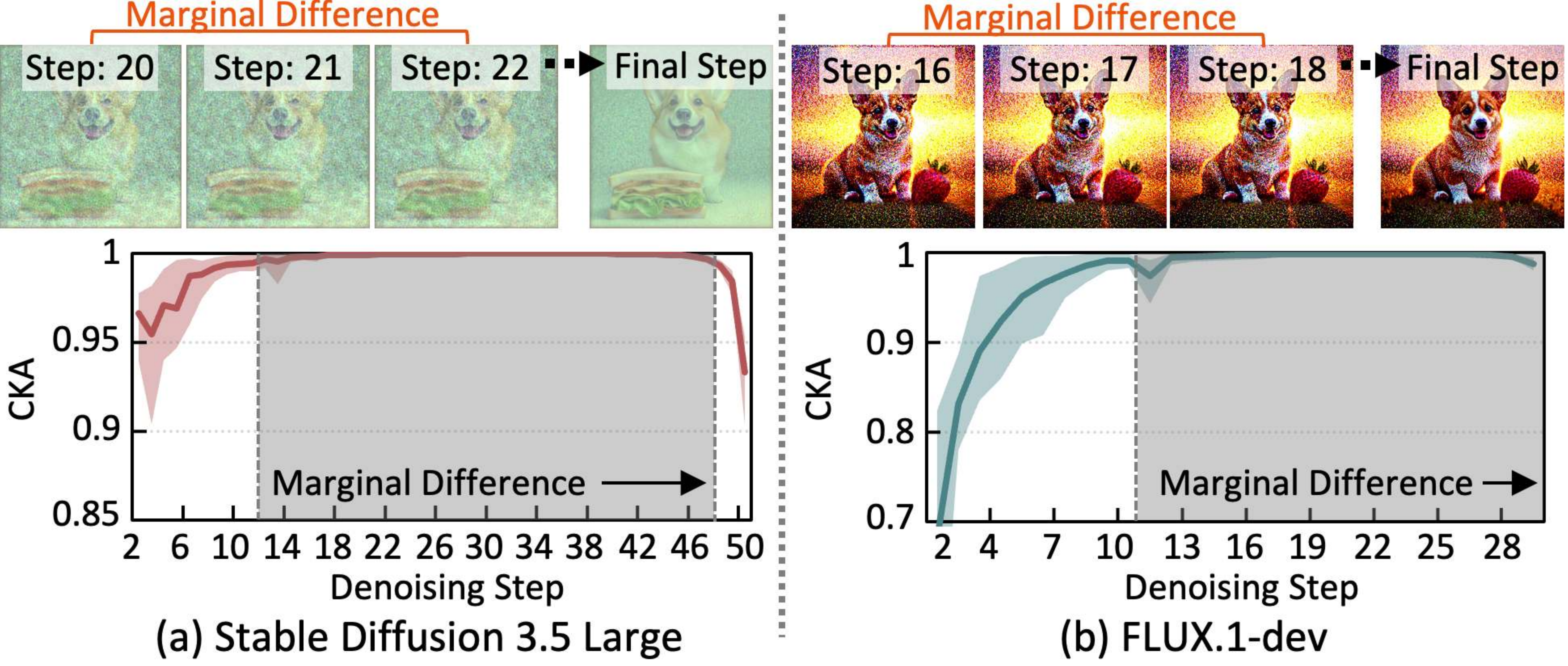}
\caption{\textbf{Similarity and redundancy across adjacent denoising steps.} CKA between the predicted noises of the adjacent denoising steps of (a) SD3.5 and (b) FLUX.1. The representations show only marginal difference across steps, as highlighted in gray-shaded area. Note the colored area represents the range across different prompts, and the colored line represents their mean value.}
\label{fig:background1}   
\end{figure}

\section{Background}

\subsection{Diffusion Models}

\paragraph{Forward and Reverse Process of Diffusion: }
The diffusion model consists of the forward process and the reverse process. In the forward process, Gaussian noise is gradually added to the real data at each timestep. In the reverse process (i.e., denoising process), diffusion models generate samples by reversing the forward process. Starting from Gaussian noise $\mathbf{x}_T \sim \mathcal{N}(0, \mathbf{I})$, the model estimates the added noise $\epsilon_\theta(\mathbf{x}_t, t)$ through the noise prediction network $\epsilon_\theta$, where $\mathbf{x}_t$ denotes the input sample at timestep $t$. The conditional probability can be modeled as in (\ref{eq:denoise}), where $\alpha_t=1-\beta_t$, $\bar{\alpha}_t=\prod_{i=1}^t \alpha_i$, and $\beta_i \in (0,1)$ denotes the noise variance schedule.
\begin{equation}
    \begin{split}
        p_\theta& (\mathbf{x}_{t-1}|\mathbf{x}_t) = \\
        &\mathcal{N}\!\left( 
        \mathbf{x}_{t-1}; 
        \frac{1}{\sqrt{\alpha_t}}
        \left(\mathbf{x}_t - 
        \frac{1-\alpha_t}{\sqrt{1-\bar{\alpha}_t}} 
        \epsilon_\theta(\mathbf{x}_t, t)\right), 
        \beta_t \mathbf{I} 
        \right).
    \end{split}
    \label{eq:denoise}
\end{equation}
During inference, this process is applied iteratively to the initial noise, producing $\mathbf{x}_0$ that approximates the data distribution.
\vspace{-10pt}

\paragraph{Diffusion Transformers: }
Recent diffusion models have increasingly adopted a transformer architecture (i.e., DiT). The noise prediction network consists of stacked transformer blocks: each with an attention block (ATTN) that mixes information across tokens and a feed-forward network block (FFN) that refines features produced by ATTN~\cite{kobayashi2024analyzing, kobayashi-etal-2020-attention, kobayashi-etal-2021-incorporating}. At each denoising step, the blocks are executed sequentially, each consuming the previous block's output. This structure is highly scalable and offers strong representational capacity, enabling larger models with better generation quality~\cite{peebles2023scalable}. However, when combined with the iterative denoising process, this larger capacity leads to higher inference cost, limiting real-time deployment.

\begin{figure}[t!]
\includegraphics[width=\linewidth]{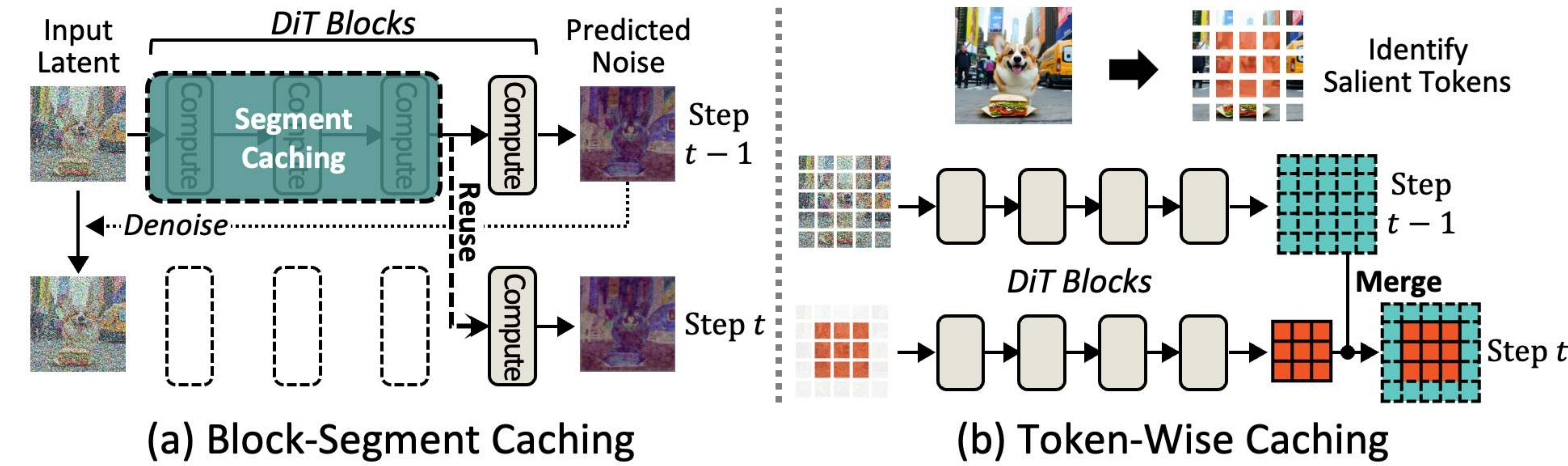}
\caption{\textbf{Overview of feature caching methods for DiT}. (a) Block-segment caching, which reuses previous step's block segment. (b) Token-wise caching, which computes salient tokens and merges them with cached output. }
\label{fig:background2}   
\end{figure}

\subsection{Redundant Computation in DiT}
\label{sec:motivation1}
Recent works have demonstrated that the iterative inference process of DiT incurs substantial redundant computation across steps~\cite{sun2024unveiling, zhang2025blockdance}. To quantify this redundancy, we analyze the representation changes between adjacent denoising steps by referring to previous works~\cite{zhang2025blockdance, li2024smartfrz, jiang2024tracing}. Fig.~\ref{fig:background1} shows CKA between adjacent steps for (a) SD3.5 and (b) FLUX.1. As shown in Fig.~\ref{fig:background1}, after the initial few steps, adjacent steps exhibit only minor representational changes, indicating high temporal redundancy--- this can also be observed in the visualized noise examples above the plot. This result reveals a potential room for inference cost reduction by removing the redundancy in the denoising process.

\subsection{Feature Caching Methods for DiT}
\label{sec:background3}
To remove the redundancy in denoising process, caching methods that reuse intermediate features have been recently proposed. Fig.~\ref{fig:background2} provides an overview of two representative caching methods: (a) block-segment caching and (b) token-wise caching.
\vspace{-10pt}

\paragraph{Block-Segment Caching: }
Fig.~\ref{fig:background2}(a) shows block-segment caching, which groups consecutive blocks into segments~\cite{chen2406delta, zhang2025blockdance, ma2024deepcache}. At each step $t$, selected segments reuse the outputs computed at the previous step $t-1$ instead of recomputing new outputs. Although such coarse-grained caching methods can improve the inference efficiency, they often skip salient information beyond the redundancy resulting in a significant quality degradation --- the output is often inconsistent with that of the original model as well. \vspace{-10pt}

\paragraph{Token-Wise Caching: }
Fig.~\ref{fig:background2}(b) shows token-wise caching, which identifies salient tokens~\cite{liu2502region,zou2024toca,cheng2025cat}. At each step $t$, the model updates only salient tokens and merges these updates with cached outputs from $t-1$ for the remaining tokens. However, consistently skipping computation for tokens that are not selected as salient can cause local quality degradation and even generate outputs that are inconsistent with those of the original model. In addition, this method often fails to capture the changes in salient tokens across blocks.

\begin{figure}[t!]
\includegraphics[width=\linewidth]{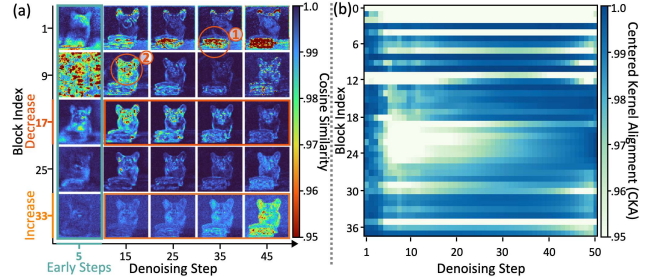}
\caption{\textbf{Block-wise contribution analysis in denoising process.} (a) Token-wise cosine similarity between the outputs from the original model and the block-pruned model of SD3.5. (b) CKA on the same pairs, evaluated across all blocks and steps}
\label{fig:motivation2}   
\end{figure}

\section{Methodology}

In this section, we first analyze DiT at the level of individual transformer blocks to accurately capture redundant computation in the denoising process (Sec.~\ref{sec:motivation3}). Motivated by the analysis, we introduce a block-wise caching approach with interval caching and intra-block caching strategy (Sec.~\ref{sec:motivation4}). We then propose a contribution-guided block-wise interval caching method, CorGi (Sec.~\ref{sec:methodology1}) together with its token-aware extension CorGi+ (Sec.~\ref{sec:methodology2}).

\subsection{Block-Wise Contribution Analysis}
\label{sec:motivation3}

Existing caching methods can degrade quality, as their coarse granularity often overlooks salient information (Sec.~\ref{sec:background3}). To better identify the source of redundant computation, we adopt a finer block-level approach by analyzing each block's role throughout the denoising process. Specifically, we quantify the per-step contribution of each block by pruning a block and analyzing how the output changes from the original model output at every denoising step using SD3.5. Fig.~\ref{fig:motivation2}(a) quantifies the contribution of each block by calculating token-wise cosine similarity of the outputs between the pruned and original execution at each step. As shown in Fig.~\ref{fig:motivation2}(a), each block is specialized for distinct regions or objects of the image. For example, block index~\#1 contributes to denoise on corgi (\circled{1}), whereas block index~\#9 contributes to denoise on the sandwich (\circled{2}). We also observe that block specialization pattern varies across denoising steps: contribution decreases in the red box, whereas it increases in the orange box as step progresses. \textbf{This observation suggests that to accurately capture the redundancy, it is crucial to consider spatially specialized and temporally varying block contributions}. 

To effectively visualize representational changes across denoising steps, we further quantify step-by-step CKA for each block in Fig.~\ref{fig:motivation2}(b) --- lower CKA denotes a greater representational change and thus a larger contribution. The CKA result further confirms that each block’s contribution varies across denoising steps, which is also aligned with our observation in Fig.~\ref{fig:motivation2}(a). Moreover, it also demonstrates that only a subset of blocks is highly active at any given step.

\begin{figure}[t!]
\includegraphics[width=\linewidth]{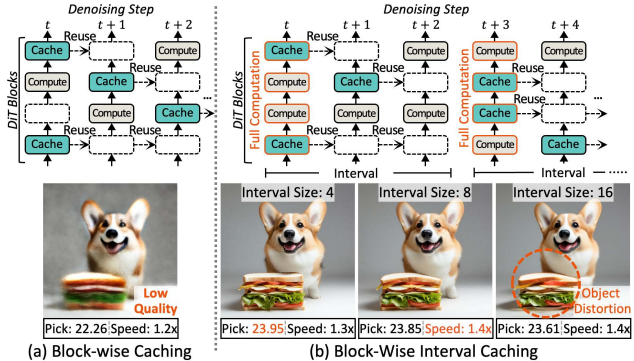}
\caption{
\textbf{Overview of block-wise caching.} (a) Naive block-wise caching which caches low-contributed blocks at each step for reuse. (b) Block-wise interval caching which refreshes cached features through full computation at intervals.
}
\label{fig:motivation3}   
\end{figure}

\paragraph{Caching Interval: }
Fig.~\ref{fig:motivation3}(a) shows the naive per-step block-wise caching method: at every step, contributions are computed via CKA, and the low-contribution blocks reuse their outputs from the previous step\footnote{In this experiment, we cache the bottom 50\% of blocks by contribution to clearly observe the effect of interval caching.}. This enables fine-grained, per-step reduction of computation by skipping updates to blocks that exhibit negligible change at each step. However, this policy often leaves some blocks persistently cached, since certain blocks maintain low contribution throughout the denoising process, as observed in Fig.~\ref{fig:motivation2}(a). This degrades the image quality. In addition, per-step calculation of the block contribution further incurs overhead, limiting the speedup gain. To overcome these limitations, we adopt interval caching, which periodically refreshes the features of all blocks at a fixed interval. At the start of each interval, we compute per-block contributions once; within the interval, the low-contribution blocks are cached and reuse their previous outputs. As shown in Fig~\ref{fig:motivation3}(b), block-wise interval caching with interval size of 4 and 8 significantly improves generation quality while achieving higher speedup. This result highlights the importance of periodic cache refresh. 
In addition, for short intervals, pre-computed contributions can be reused throughout the interval with minimal loss --- block contributions exhibit minor changes over short durations, as observed in Fig.~\ref{fig:motivation2}(b). Note more ablation details are presented in Sec.~\ref{sec:experiment2}.

\subsection{Block-Wise Caching}
\label{sec:motivation4}
Motivated by the observations in Sec.~\ref{sec:motivation3}, we first propose a block-wise caching method that selectively skips computations of low-contribution blocks by reusing cached outputs from previous steps. To adequately capture the temporally varying block contributions while preserving salient information accumulated from preceding blocks, we explore two important design components: caching interval and intra-block caching strategy.
\vspace{-10pt}

\begin{figure}[t!]
\includegraphics[width=\linewidth]{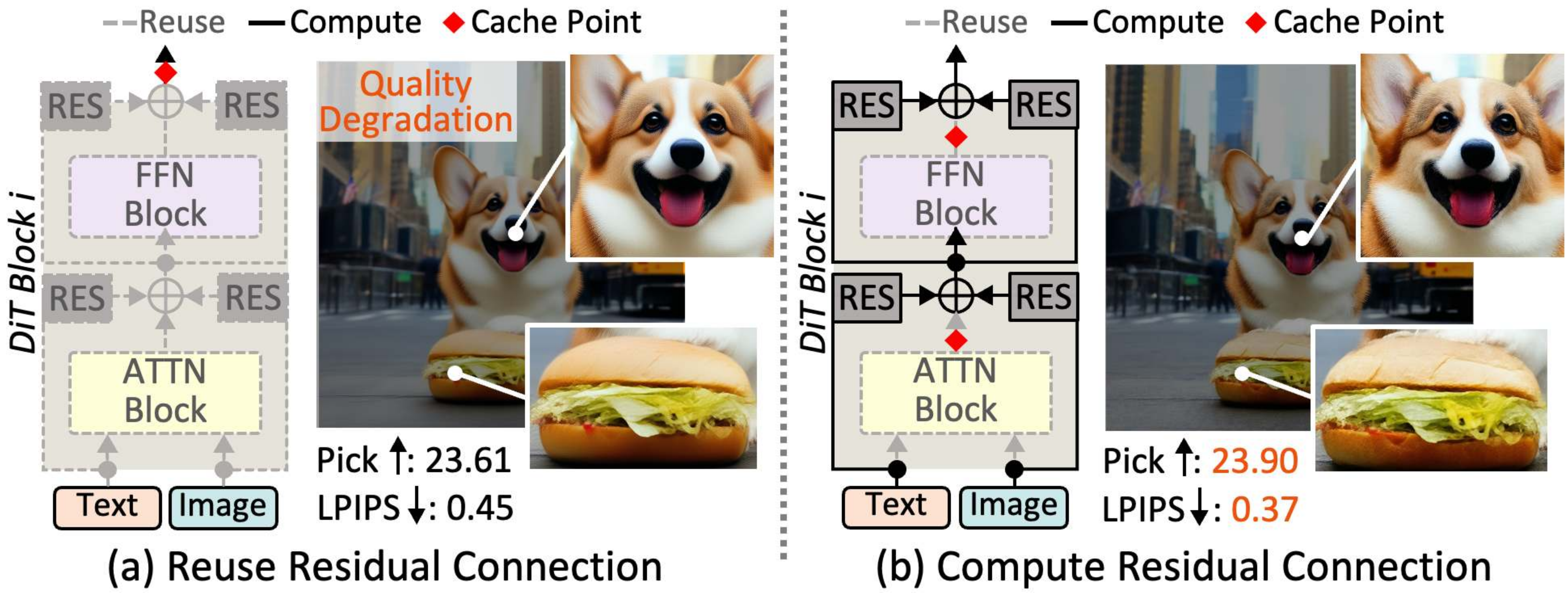}
\caption{
\textbf{Overview of intra-block caching strategy}. (a) Caching the entire block output without updating the residual connection. (b) Caching the ATTN and FFN outputs while updating the residual connection.}
\label{fig:motivation4}  

\end{figure}

\paragraph{Intra-Block Caching Strategy: }
Given the stacked, sequential execution of DiT blocks, naively reusing block outputs can overwrite the outputs of previous blocks as they propagate through the stack~\cite{chen2406delta}. This leads to information vanishing along the block stack, resulting in quality degradation as shown in Fig.~\ref{fig:motivation4}(a). To avoid this, we do not cache the entire block output. Instead, we cache the outputs of the ATTN and FFN modules while continuously updating the residual connection of every block at each step. This preserves the accumulated contributions from preceding blocks. As shown in Fig.~\ref{fig:motivation4}(b), this intra-block caching strategy maintains fine details in the generated images by reusing ATTN and FFN outputs while continuously updating the residual connection.

\begin{figure*}[t!]
\includegraphics[width=\linewidth]{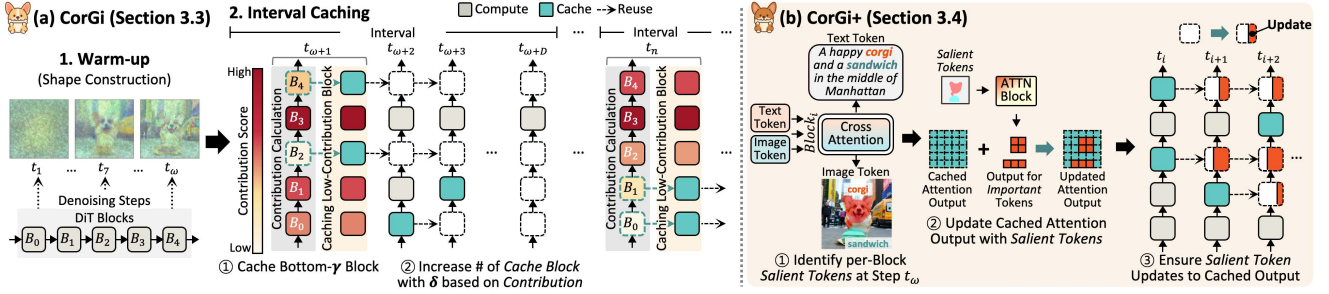}
\caption{
\textbf{Design overview of CorGi and CorGi+.}
(a) CorGi computes all DiT blocks for the first $\omega$ denoising steps and gradually caches the outputs of low-contribution blocks and reuses them in later steps within interval guided by contribution score.
(b) CorGi+ identifies salient tokens per block by leveraging cross-attention maps and ensures partial updates for the salient tokens.
}
\label{fig:overview}   
\end{figure*}

\subsection{CorGi: Contribution-Guided Block-Wise\\Interval Caching}
\label{sec:methodology1}
Building on our analysis, we propose a contribution-guided block-wise interval caching method, \textbf{CorGi} (Fig.~\ref{fig:overview}(a)). CorGi consists of two phases: warm-up and interval caching. As early steps are critical for constructing the image's global structure (Fig.~\ref{fig:motivation2}(a)), CorGi begins with a warm-up phase to ensure stable image generation. In the warm-up phase, CorGi computes all DiT blocks for $\omega$ \footnote{In our experiments, we empirically set $\omega$ to 20\% of the total denoising steps. Larger $\omega$ improves generation quality but reduces speed gain, whereas smaller $\omega$ yields higher speedup at the risk of quality degradation.} denoising steps. After the $\omega$ steps, CorGi moves into interval caching phase. In each interval, CorGi calculates contributions of blocks. Based on the calculated contributions, CorGi caches low-contributing blocks. Details of the interval caching phase are as follows. \vspace{-10pt}

\paragraph{Contribution Calculation: }
To quantify the contribution of each block, CorGi measures how much each block's feature representation changes between two consecutive intervals as denoising progresses, using CKA --- similar practice has been used to capture representation changes in previous works~\cite{son2025safe, neyshabur2020being, raghu2021vision}. Accordingly, CorGi computes CKA between the outputs of block $i$ at the initial step of two consecutive intervals as:
\begin{equation} 
\label{eq:CKA}
\text{CKA}_{i}(X_i, Y_i) = \frac{\|Y_i^T X_i\|_F^2}{\|X_i^T X_i\|_F \|Y_i^T Y_i\|_F},
\end{equation} 

\noindent where $X_i$ and $Y_i$ denote the feature representation of block $i$ at the initial step of the current and the previous interval, respectively. A higher CKA value indicates that the block's feature has not much changed as denoising progresses within interval, suggesting limited contribution on the denoising process. To this end, CorGi calculates the contribution score of block $i$ as $1 - \text{CKA}_{i}(X_i, Y_i)$.

\paragraph{Block-Wise Interval Caching: }
At the initial step of each interval (which is composed of $D$ steps), CorGi performs full computation over all blocks to obtain updated outputs. It then calculates each block's contribution score by comparing the updated outputs with the initial step outputs from the previous interval. Within an interval, CorGi selects the $\gamma$ low-contribution blocks as the initial caching targets. CorGi then caches the outputs of the selected blocks so that the cached outputs can be reused in later steps of the interval. 

To further reduce redundant computation, CorGi gradually increases $\delta$ additional caching blocks as the step proceeds within interval, based on the contribution scores computed at the initial step of current interval. This design choice ensures full computation for high-contribution blocks within the interval, while the amount of computations is reduced for lower contribution blocks according to the contribution score. By doing so, CorGi can achieve higher caching ratio while not much degrading the generation quality --- the detailed quality-speedup trade-off analysis is presented in Sec.~\ref{sec:experiment2}.

\subsection{CorGi+: Salient Token Protection}
\label{sec:methodology2}
As different blocks concentrate on different objects or spatial regions (Sec.~\ref{sec:motivation3}), the set of active tokens (i.e., \textit{salient tokens}) within each block also varies across blocks. To identify these block-specific salient tokens, we leverage cross-attention maps in T2I diffusion models, which reveal token-level image–text associations, enabling the identification of tokens associated with semantically important objects~\cite{kimseg4diff, shin2025exploring, hertz2022prompt}. To this end, we propose \textbf{CorGi+} (Fig.~\ref{fig:overview}(b)) which leverages per-block cross-attention to select salient tokens for each block and ensures partial updates for the selected salient tokens to enhance generation quality.

\vspace{-10pt}

\paragraph{Salient Token Identification: }
After the warm-up phase has established the global structure, CorGi+ computes salient tokens from each block’s cross-attention maps. 
Let $A_i \in \mathbb{R}^{L_{\text{img}}\times L_{\text{text}}}$ denote the cross-attention map of block $i$ at step $\omega$, where $L_{\text{img}}$ and $L_{\text{text}}$ denote the numbers of image and text tokens, respectively. 
For each text token $u \in e^{\text{text}}$ and image token $v \in e^{\text{img}}$,
where $e^{\text{text}} \in \mathbb{R}^{L_{\text{text}}}$ and $e^{\text{img}}  \in \mathbb{R}^{L_{\text{img}}}$ represent the sets of text and image tokens respectively, we define its saliency score $\rho$ as:
\begin{equation}
\label{eq:saliency_score}
\rho_{i,u} \;=\; \max_{v\; \in \;
e^{\text{img}}
} A_i(v,u).
\end{equation}

CorGi+ then selects the top-$c$\footnote{We determine $c$ for each model from empirical evaluation.} text tokens per block by $\rho_{i,u}$ to obtain the salient text token set as:
\begin{equation}
\label{eq:top_c_text}
S_i^{\text{text}}
= \bigl\{\, u \;\big|\; \rho_{i,u} \in \operatorname{Top}\!-\!c\; \big(
\{\,\rho_{i,u} \mid u \in e^{\text{text}}\,\}
\big) \bigr\}
\end{equation}

For each salient text token $u \in S^{\text{text}}_{i}$, we perform $k$-means clustering\footnote{We set $k=2$ based on empirical results across models.} over the image-token scores in the column vector $A_i(:,u)$ and take the high-attention cluster as the set of image tokens strongly related to $u$ as:
\begin{equation}
\label{eq:kmeans_img}
S^{\text{img}}_{i}(u) \;=\; \operatorname{HighCluster}\big(\operatorname{k\text{-}means}(A_i(:,u),\,k)\big).
\end{equation}

This procedure yields, for each block $i$, a compact set of \textit{salient tokens} 
\(
S_{i} = S^{\text{text}}_{i} \cup S^{\text{img}}_{i}
\)
comprising both modalities—the top-$c$ salient text tokens $S^{\text{text}}_{i}$ and,
for each $u \in S^{\text{text}}_{i}$, the corresponding salient image tokens aggregated as
\(S^{\text{img}}_{i}=\bigcup_{u\in S^{\text{text}}_{i}} S^{\text{img}}_{i}(u)\).
CorGi+ later ensures to update all tokens in $S_{i}$ even if the block uses cached outputs.
\vspace{-10pt}

\begin{table}[t!]
\centering
\resizebox{\columnwidth}{!}{
\begin{tabular}{l | c | cccc}
\toprule
\textbf{Method} & Speedup & FID ↓ & Pick ↑ & IR ↑ & LPIPS ↓ \\
\midrule
SD 3.5 Large, 50 steps & \textcolor{customblue}{--} & 20.01 & 22.90 & 1.09 & -- \\
\midrule
DeepCache ($N$=2)& \textcolor{customblue}{\textbf{1.29$\times$}} & 19.44 & 22.85 & 1.07 & 0.24 \\
$\Delta$-DiT ($b$=30) & \textcolor{customblue}{\textbf{1.62$\times$}} & 19.72 & 22.85 & 1.06 & 0.25 \\
BlockDance ($N$=3) & \textcolor{customblue}{\textbf{1.49$\times$}} & 20.26 & 22.89 & 1.08 & 0.04 \\
RAS (30\% ratio) & \textcolor{customblue}{\textbf{1.99$\times$}} & 21.50 & 22.45 & 1.00 & 0.11 \\
\midrule
\rowcolor[gray]{0.95}
CorGi ($D$=6, $\gamma$=28, $\delta$=2) & \textcolor{customblue}{\textbf{2.08$\times$}} & 21.45 & 22.73 & 1.02 & 0.09 \\
\rowcolor[gray]{0.95}
CorGi ($D$=5, $\gamma$=30, $\delta$=2) & \textcolor{customblue}{\textbf{2.09$\times$}} & 21.38 & 22.74 & 1.02 & 0.08 \\

\midrule
\rowcolor[gray]{0.9}
CorGi+ ($D$=6, $\gamma$=28, $\delta$=2) & \textcolor{customblue}{\textbf{2.02$\times$}} & 20.71 & 22.74 & 1.04 & 0.09 \\
\rowcolor[gray]{0.9}
CorGi+ ($D$=5, $\gamma$=30, $\delta$=2) & \textcolor{customblue}{\textbf{2.02$\times$}} & 20.70 & 22.76 & 1.05 & 0.08 \\

\bottomrule
\end{tabular}
}
\caption{\textbf{Text-to-image generation comparison on Stable Diffusion 3.5 Large.}}
\label{tab:sdL}
\end{table}

\paragraph{Ensuring Salient Token Updates: }
For all blocks that use cached output, CorGi+ partially updates the block-specific salient tokens in ATTN, and merges them into the cached output. We restrict these partial updates to ATTN because attention activations align closely with object- and region-level semantics in T2I~\cite{kimseg4diff, shin2025exploring, hertz2022prompt}, thereby minimizing overhead while preserving quality.

Let $S_i \subseteq e^{\text{text}} \cup e^{\text{img}}$ denote the salient token set defined earlier. Based on this set, we construct a binary mask $M_i \in \{0,1\}^{L_{\text{text}} + L_{\text{img}}}$ such that $[M_i]_u=1$ if and only if $u \in S_i$. We denote the cached ATTN outputs by $\hat{Z}^{\text{attn}}_{i}$ and the updated ATTN output at step $t$ by $\widetilde{Z}^{\text{attn}}_{i}(t)$, which is computed only on the tokens contained in $S_i$.

To compose block outputs, CorGi+ performs masked replacement for ATTN only on $S_i$ as:
\begin{equation}
\label{eq:masked_attn_merge}
Z^{\text{attn}}_{i}(t) \;=\; M_i \odot \widetilde{Z}^{\text{attn}}_{i}(t) \;+\; (1-M_i)\odot \hat{Z}^{\text{attn}}_{i},
\end{equation}
where $\odot$ denotes elementwise multiplication with broadcasting over feature dimensions.

Since $|S_i| \ll L_{\text{text}}+L_{\text{img}}$, the fraction of updated tokens in ATTN remains low, yielding minimal overhead while preserving fine details in every block.

\begin{table}[t!]
\centering
\resizebox{\columnwidth}{!}{
\begin{tabular}{l | c | cccc}
\toprule
\textbf{Method} & Speedup & FID ↓ & Pick ↑ & IR ↑ & LPIPS ↓ \\
\midrule
FLUX.1-dev, 30 steps & \textcolor{customblue}{--} & 27.77 & 23.11 & 1.08 & -- \\
\midrule
DeepCache ($N$=2) & \textcolor{customblue}{\textbf{1.26$\times$}} & 26.24 & 23.11 & 1.09 & 0.25 \\
BlockDance ($N$=3)& \textcolor{customblue}{\textbf{1.50$\times$}} & 26.24 & 23.12 & 1.10 & 0.05 \\
RAS (50\% ratio)& \textcolor{customblue}{\textbf{1.98$\times$}} & 28.01 & 22.76 & 1.01 & 0.14 \\
\midrule
\rowcolor[gray]{0.95}
CorGi ($D$=5, $\gamma$=49, $\delta$=2) & \textcolor{customblue}{\textbf{2.00$\times$}} & 28.57 & 22.97 & 1.06 & 0.09 \\
\rowcolor[gray]{0.95}
CorGi ($D$=6, $\gamma$=47, $\delta$=2) & \textcolor{customblue}{\textbf{2.08$\times$}} & 28.05 & 22.87 & 1.05 & 0.10 \\
\midrule
\rowcolor[gray]{0.9}
CorGi+ ($D$=5, $\gamma$=49, $\delta$=2) & \textcolor{customblue}{\textbf{1.91$\times$}} & 27.56 & 22.95 & 1.05 & 0.09 \\
\rowcolor[gray]{0.9}
CorGi+ ($D$=6, $\gamma$=47, $\delta$=2) & \textcolor{customblue}{\textbf{1.98$\times$}} & 27.48 & 22.82 & 1.01 & 0.12 \\

\bottomrule
\end{tabular}
}
\caption{\textbf{Text-to-image generation comparison on FLUX.1-dev.}}
\label{tab:flux}
\end{table}

\begin{table}[t!]
\resizebox{\columnwidth}{!}{%
\centering
\begin{tabular}{l | c | cccc}
\toprule
\textbf{Method} & Speedup & FID ↓ & Pick ↑ & IR ↑ & LPIPS ↓ \\
\midrule
PixArt-$\Sigma$, 50 steps & \textcolor{customblue}{--} & 29.94 & 22.69 & 0.87 & -- \\
\midrule
DeepCache ($N$=2) & \textcolor{customblue}{\textbf{1.31$\times$}} & 30.38 & 22.62 & 0.82 & 0.13 \\
$\Delta$-DiT ($b$=30) & \textcolor{customblue}{\textbf{1.59$\times$}} & 28.77 & 22.73 & 0.89 & 0.11 \\
BlockDance ($N$=3)& \textcolor{customblue}{\textbf{1.54$\times$}} & 35.91 & 22.41 & 0.72 & 0.21 \\
\midrule
\rowcolor[gray]{0.95}
CorGi ($D$=4, $\gamma$=22, $\delta$=2) & \textcolor{customblue}{\textbf{1.93$\times$}} & 27.56 & 22.60 & 0.83 & 0.15 \\
\rowcolor[gray]{0.95}
CorGi ($D$=5, $\gamma$=20, $\delta$=2) & \textcolor{customblue}{\textbf{2.00$\times$}} & 27.64 & 22.50 & 0.82 & 0.19 \\

\midrule
\rowcolor[gray]{0.9}
CorGi+ ($D$=4, $\gamma$=22, $\delta$=2) & \textcolor{customblue}{\textbf{1.91$\times$}} & 27.58 & 22.60 & 0.83 & 0.15 \\
\rowcolor[gray]{0.9}
CorGi+ ($D$=5, $\gamma$=20, $\delta$=2) & \textcolor{customblue}{\textbf{1.97$\times$}} & 27.67 & 22.50 & 0.82 & 0.19 \\
\bottomrule
\end{tabular}%
}
\caption{\textbf{Text-to-image generation comparison on PixArt-$\Sigma$.}}
\label{tab:pixart}
\end{table}

\begin{figure*}[t!]
\includegraphics[width=\linewidth]{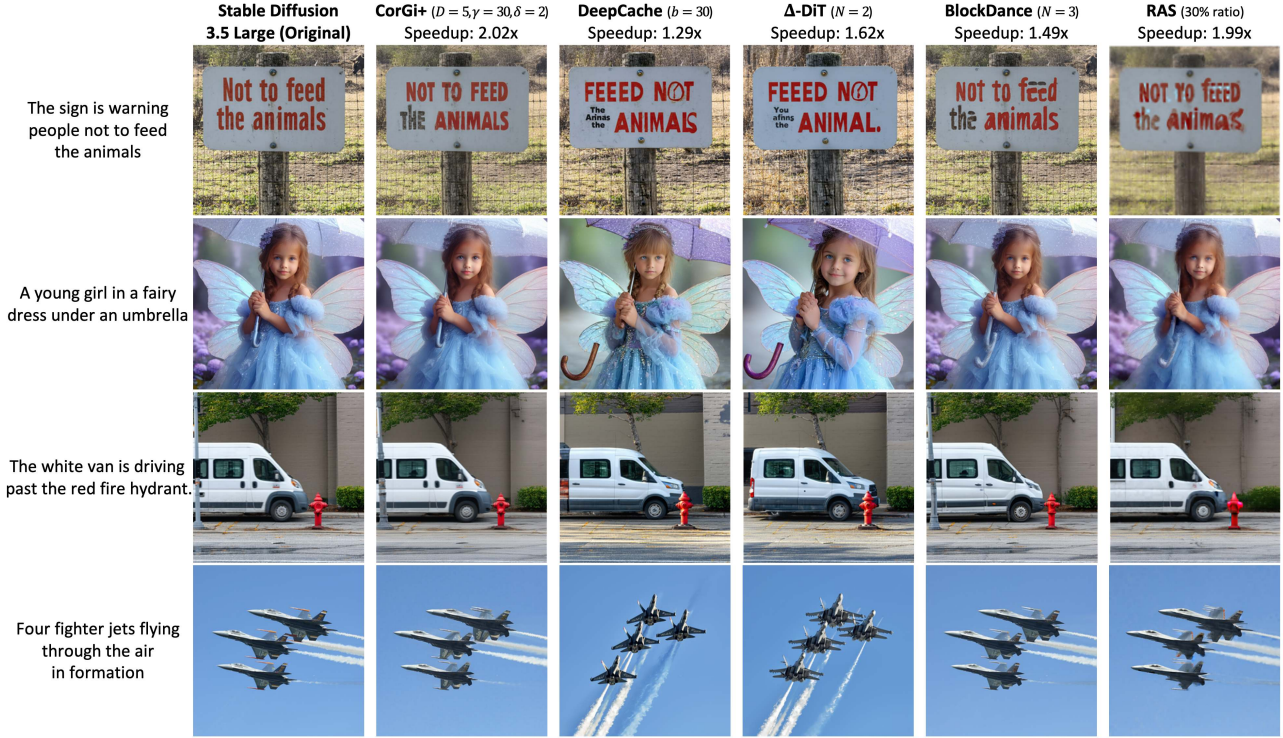}
\caption{\textbf{Qualitative comparisons.} CorGi+ achieves the highest speedup while maintaining strong consistency with the original images.}
\label{fig:result1}   
\end{figure*}

\section{Experiments}

\subsection{Experimental Settings}

\paragraph{Models, Dataset and Metrics: }
We conduct an evaluation on text-to-image generation across three DiT models: Stable Diffusion 3.5 Large~\cite{SD3.5}, FLUX.1-dev~\cite{FLUX}, and PixArt-$\Sigma$~\cite{chen2024pixart}. For all experiments, we generate 1024$\times$1024 images on the 10K prompts from the validation set of MS-COCO 2014~\cite{lin2014microsoft}, with a guidance scale of 3.5. To assess generation quality, we report FID~\cite{heusel2017gans}, PickScore~\cite{kirstain2023pick} and ImageReward~\cite{xu2023imagereward}. In addition, we use LPIPS~\cite{zhang2018unreasonable} to measure the similarity between the outputs of the accelerated models and those of the original models. We measure the end-to-end inference speedup on an NVIDIA H100 GPU. 
\vspace{-10pt}

\paragraph{Baselines: }
We evaluate CorGi and CorGi+ against state-of-the-art block-segment and token-wise caching methods. For block-segment caching methods, we include DeepCache~\cite{ma2024deepcache}, $\Delta$-DiT~\cite{chen2406delta}, and BlockDance~\cite{zhang2025blockdance}. For a token-wise caching method, we include RAS~\cite{liu2502region}. The detailed parameter settings of our proposed methods and baselines are summarized in Tables~\ref{tab:sdL}, \ref{tab:flux}, and \ref{tab:pixart}.

\subsection{Main Result}
\label{sec:experiment1}

\paragraph{Accelerating Text-to-Image Generation: }

Table~\ref{tab:sdL} reports the results on Stable Diffusion 3.5 Large. CorGi achieves substantial inference speedups by skipping low-contribution blocks at each step. At the same time, it maintains image quality comparable to the original model and attains lower LPIPS, indicating high consistency with minimal quality degradation. CorGi+ further improves over CorGi with minor computational overhead, achieving higher speedups and lower LPIPS while better preserving object details.

Compared to DeepCache and $\Delta$-DiT, CorGi+ attains comparable image quality while achieving higher speedups and lower LPIPS. This is because both baselines apply caching from the initial step without a warm-up phase, which can distort image structure and reduce consistency with the original model outputs. Compared to BlockDance, CorGi+ attains comparable image quality while achieving a 26.2\% higher speedup, as BlockDance’s segment-level caching remains too coarse-grained to effectively remove redundant computation. CorGi+ outperforms RAS across all evaluation metrics, as RAS prioritizes updates only on salient tokens, resulting in quality degradation. In summary, CorGi and CorGi+ efficiently accelerate DiT inference while preserving generation quality, achieving a better speed–quality trade-off compared to the baselines.

We observe similar trends on FLUX.1, as reported in Table~\ref{tab:flux}. These results indicate that CorGi and CorGi+ can effectively accelerate DiT inference regardless of the underlying model architecture, making them broadly applicable across diverse model architectures and scales.

Table~\ref{tab:pixart} reports the results on PixArt-$\Sigma$. CorGi consistently outperforms all baselines in FID and even surpasses the original model. While maintaining comparable performance on the other metrics, CorGi achieves up to 2.00$\times$ inference speedup. In contrast, CorGi+ provides limited additional gains on PixArt-$\Sigma$, mainly due to the relatively small model size and the limited ability of PixArt-$\Sigma$ to exploit text–image relationships for salient-token identification.

\vspace{-10pt}

\paragraph{Qualitative Comparisons: }
Fig.~\ref{fig:result1} further presents a qualitative comparison against the baselines. DeepCache and $\Delta$-DiT often generate high-quality images, but the results are semantically inconsistent with the original model outputs. BlockDance generally produces high-quality images, but its coarse segment-level caching results in only limited speedup. RAS often generates images with degraded details and noisy textures. In contrast, CorGi+ achieves the highest speedup (2.02$\times$) while maintaining high visual consistency with the original model outputs and preserving fine-grained image details. Note additional qualitative results for CorGi and CorGi+ are presented in Appendix B.

\begin{figure}[t!]
\includegraphics[width=\linewidth]{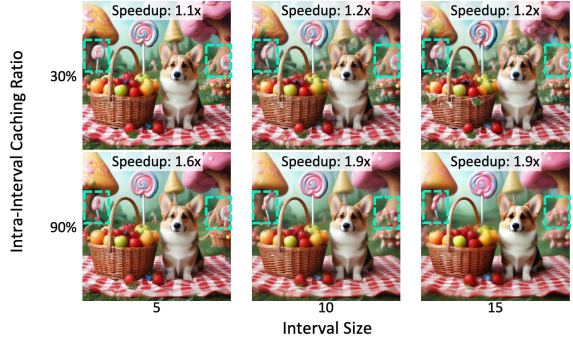}
\caption{\textbf{Sensitivity analysis.} The highlighted regions demonstrate that increasing either the interval size or the intra-interval caching ratio degrades object details despite the speed gains.}
\label{fig:result2}   
\end{figure}

\subsection{Ablation Study}
\label{sec:experiment2}

\paragraph{Caching Granularity: }

We study the caching granularity of CorGi to the interval size $D$ and the intra-interval caching ratio controlled by $\gamma$ and $\delta$. As shown in Fig.~\ref{fig:result2}, increasing $D$ makes cached blocks stay unchanged for longer, which can cause object structures to become distorted or even disappear. This can lead to a noticeable loss of fine-grained details despite higher speedup. We also observe that a higher intra-interval caching ratio (larger $\gamma$ and $\delta$) achieves greater speedup but simultaneously degrades image quality. This reveals a clear accuracy–speedup tradeoff, highlighting the importance of carefully choosing $D$, $\gamma$, and $\delta$ to balance efficiency and generation quality.
\vspace{-10pt}

\paragraph{Impact of Salient Token Protection: }

\begin{figure}[t!]
\includegraphics[width=\linewidth]{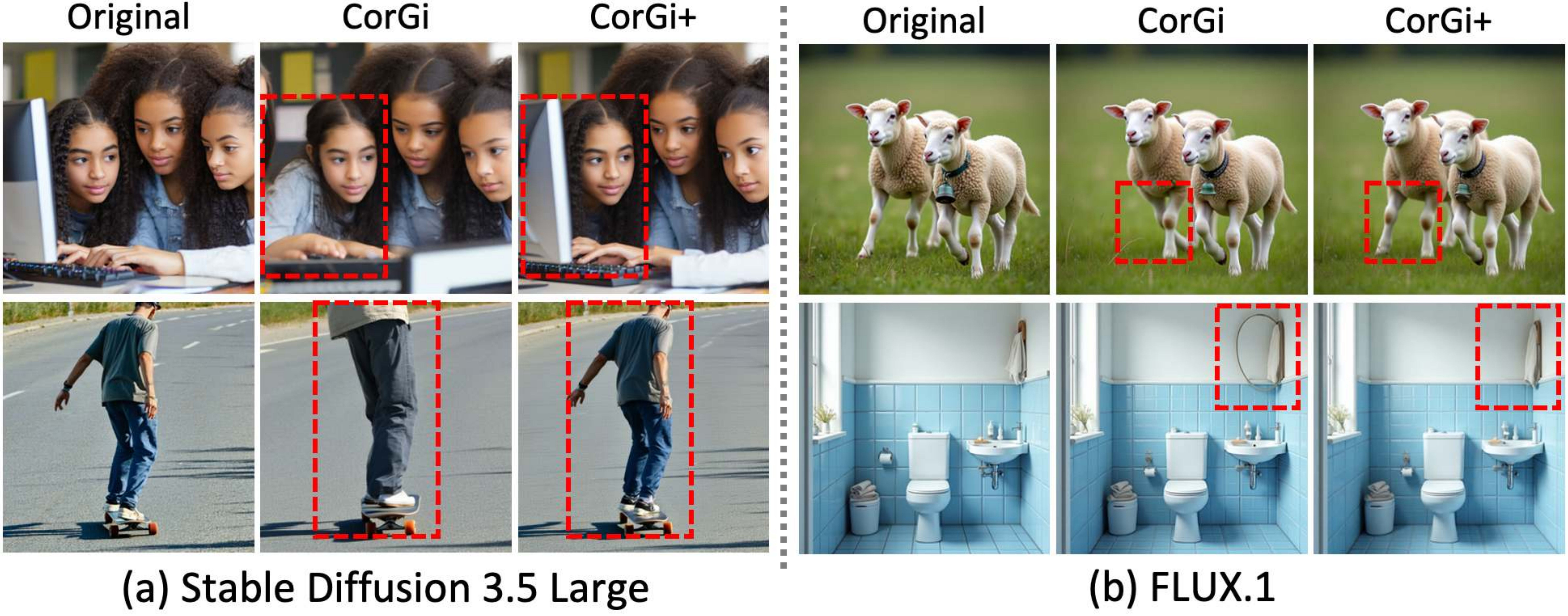}
\caption{\textbf{Ablation on CorGi+.} The highlighted regions demonstrate that CorGi+ restores degraded object details to closely match the original images for both (a) SD3.5 and (b) FLUX.1.}
\label{fig:result3}   
\end{figure}

We investigate the impact of salient token protection in CorGi+. As shown in Fig.~\ref{fig:result3}, CorGi+ effectively identifies the salient tokens associated with the red highlighted regions and restores the corresponding objects for both (a) SD3.5 and (b) FLUX. For example, in the top example of Fig.~\ref{fig:result3}(a), the monitor region becomes severely degraded under CorGi, whereas CorGi+ preserves the salient tokens associated with the monitor and successfully restores its shape and details to closely match the original image. These results indicate that CorGi+ successfully identifies salient tokens within each block and guarantees their updates, thereby preserving fine-grained object details.

\section{Previous Work}
Existing diffusion acceleration methods can be categorized into non-caching and caching approaches. Non-caching methods accelerate denoising by reducing the number of denoising steps while maintaining generation quality. DDIM~\cite{song2020denoising} achieves faster sampling by replacing the stochastic DDPM process~\cite{ho2020denoising} with a non-Markovian process which allows fewer denoising steps. DPM-Solver~\cite{lu2022dpm} formulates the denoising process as an ODE and introduces efficient high-order solvers. Step-distillation methods~\cite{salimans2022progressive, meng2023distillation} train a student model to approximate the denoising process in fewer iterations. Our approach can be applied atop the above methods to further remove block-level redundancy for faster denoising.

Caching-based approaches accelerate the denoising process by reusing intermediate features across denoising steps to reduce redundant computation. Block-segment caching groups consecutive blocks and reuses their outputs across steps.
$\Delta$-DiT~\cite{chen2406delta} characterizes the roles of front and rear parts of DiT blocks and leverages this for caching. BlockDance~\cite{zhang2025blockdance} exploits the role of front parts of DiT blocks, caching them after the initial steps. Several methods apply block-segment caching within specific components of each block. T-GATE~\cite{liu2024faster} models the distinct contributions of ATTN and adaptively gates it across steps. FORA~\cite{selvaraju2024fora} caches outputs of ATTN and FFN by exploiting feature redundancy between adjacent steps. Although such caching strategies improve the inference speedup, they often overlook salient information, leading to quality degradation and inconsistency with the outputs from the original model.

As DiT typically processes all tokens equally regardless of their perceptual or semantic differences, recent works introduce token-wise caching. RAS~\cite{liu2502region} uses the local standard deviation of the predicted noise to identify the salient tokens for update. CAT Pruning~\cite{cheng2025cat} clusters image tokens and selects clusters for update based on the variation in noise. ToCa~\cite{zou2024toca} computes token importance using ATTN influence, caching frequency, and spatial redundancy, and selects salient tokens based on these scores. By prioritizing these salient tokens, such strategies can leave the remaining tokens insufficiently updated, causing local quality degradation and inconsistency with outputs of the original model.

Different from existing caching methods, CorGi and CorGi+ precisely remove redundant computation with block-level approach, thereby preserving salient information and achieving improved speedups while maintaining high generation quality.

\section{Conclusion}

In this paper, we propose CorGi, a training-free, contribution-guided block-wise interval caching method that removes redundant computation while preserving high generation quality. We further propose CorGi+, which preserves fine object details by updating block-specific salient tokens in attention. Our proposed methods achieve 2.0$\times$ faster inference over the state-of-the-art DiT models, while preserving high generation quality.

\bibliographystyle{ieeenat_fullname}
\bibliography{references}

\clearpage

\appendix
\clearpage
\maketitlesupplementary

In the supplementary material, Sec.~\ref{sec:appendix1} provides additional ablation studies of our proposed method. Sec.~\ref{sec:ablation2} illustrates the impact of our intra-caching strategy on generation quality. Sec.~\ref{sec:ablation3} analyzes how warm-up steps influence image quality. Sec.~\ref{sec:ablation4} illustrates the impact of our contribution scores by visualizing the block-level caching patterns across denoising steps. Sec.~\ref{sec:ablation5} explains how salient tokens are identified and how partial update preserves fine-grained object details. Sec.~\ref{sec:appendix2} presents additional qualitative results for Stable Diffusion 3.5 Large, FLUX.1-dev, and PixArt-$\Sigma$. Sec.~\ref{sec:appendix3} presents additional experiments on FLUX.1-dev using the 5K prompts from the MS COCO validation set.

\section{Additional Ablation Studies}
\label{sec:appendix1}

\subsection{Impact of Intra-Block Caching Strategy}
\label{sec:ablation2}

\paragraph{Formulation of Intra-Block Caching Strategy: }
To understand the role of residual updates within an intra-block caching strategy, we first formalize the computation of a DiT block. Let $h_{t,i}$ denote the input to block $i$ at step $t$, and let $\text{ATTN}_i(\cdot)$ and $\text{FFN}_i(\cdot)$ represent the ATTN and FFN block computations for block $i$, respectively. A DiT block computation can be expressed as

\begin{equation}
    \label{eq:block_eq}
    \begin{aligned}
        h_{t, i+1} 
        &= h_{t, i} + \text{ATTN}_i(h_{t, i}) \\
        &\quad +\text{FFN}_i(h_{t, i} + \text{ATTN}_i(h_{t, i}))
    \end{aligned}
\end{equation}

We distinguish two ways of reusing block outputs: (i) caching the entire block output, including the residual connection, and (ii) caching only the outputs of the ATTN and FFN blocks while computing the residual connection.
First, when the residual connection is reused, the block output takes the following form:

\begin{equation}
\label{eq:reusing_residual}
\begin{aligned}
\hat{h}_{t, i+1}^{\text{reuse}}
&= h_{\hat{t}, i} + \text{ATTN}_i(h_{\hat{t}, i}) \\
&\quad + \text{FFN}_i\!\left(h_{\hat{t}, i} + \text{ATTN}_i(h_{\hat{t}, i})\right)
\end{aligned}
\end{equation}
where $t > \hat{t}$ and $\hat{t}$ denotes the step at which the block output is cached.
In contrast, when the residual connection is computed while reusing only the ATTN and FFN block outputs, the block output becomes:

\begin{equation}
\label{eq:compute_residual}
\begin{aligned}
\hat{h}_{t, i+1}^{\text{compute}}
&= h_{t, i} + \text{ATTN}_i(h_{\hat{t}, i}) \\
&\quad + \text{FFN}_i\!\left(h_{\hat{t}, i} + \text{ATTN}_i(h_{\hat{t}, i})\right)
\end{aligned}
\end{equation}

\paragraph{Importance of Residual Connection Updates: }
To maintain consistency between the outputs of the cached and original models, the denoising path under caching should remain aligned with the original, non-cached path. However, as shown in Eqn.~\ref{eq:reusing_residual}, reusing the residual connection injects input information from a previous step into the current one, thereby affecting the current denoising path. Since the residual connection serves as the primary mechanism for preserving the original input information in transformer block~\cite{kobayashi-etal-2021-incorporating}, computing the residual connection allows the model to properly preserve the input information of the current step, as shown in Eqn.~\ref{eq:compute_residual}.

\begin{figure}[!t]
\centering
\includegraphics[width=\linewidth]{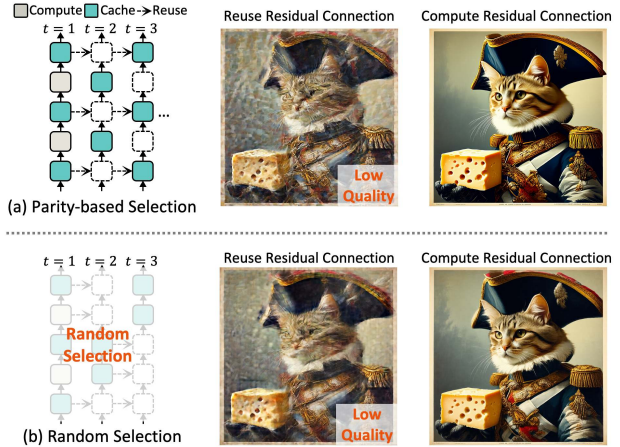}
\caption{\textbf{Ablation on intra-block caching strategy.} Comparison between the results when reusing the residual connection and computing the residual connection under (a) parity-based block selection and (b) random block selection.}
\label{fig:ablation1}   
\end{figure}

Fig.~\ref{fig:ablation1} illustrates the effectiveness of our intra-caching strategy by comparing residual reuse and residual computation across two block caching strategies. As shown in Fig.~\ref{fig:ablation1}, in parity-based block selection, where connections to the previous block are repeatedly disconnected, upstream information cannot be properly propagated. In this setting, reusing the residual connection causes the current input to be strongly affected by the previous step, leading to severe image collapse. In contrast, with our intra-block caching strategy, upstream information is correctly propagated along the denoising path, enabling stable image generation. Fig.~\ref{fig:ablation1}(b) further presents the results under random block selection. Similarly, we observe that continuously computing the residual connection preserve generation quality, whereas reusing the residual connection leads to noticeable degradation in the output.

\subsection{Impact of Warm-up}
\label{sec:ablation3}

\begin{figure}[!t]
\centering
\includegraphics[width=\linewidth]{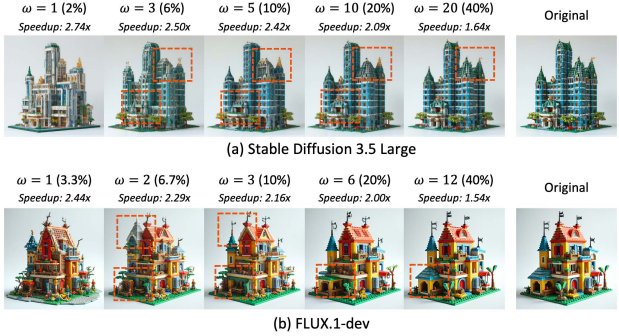}
\caption{\textbf{Ablation on warm-up.} Impact of the number of warm-up steps for (a) Stable Diffusion 3.5 Large and (b) FLUX.1-dev. }
\label{fig:ablation2}   
\vspace{-5pt}
\end{figure}

As discussed in Sec.~\ref{sec:methodology1}, early denoising steps play a crucial role in constructing the global image structure; accordingly, CorGi employs a warm-up phase of $\omega$ steps. We explore multiple warm-up ratios based on the total number of denoising steps for each model --- 50 steps for SD3.5 and 30 steps for FLUX --- and observe how these different warm-up ratios affect the generated outputs. As shown in Fig.~\ref{fig:ablation2}, when we set the warm-up length to $\omega=1$, the structure of the generated images differs significantly from that of the original outputs. As the warm-up ratio increases, the global structure and object details are progressively recovered. However, increasing the warm-up ratio also increases the number of steps with full computation, thereby reducing the overall speed gains. This highlights the inherent trade-off between acceleration and generation quality when choosing the warm-up length.

\begin{figure}[!t]
\centering
\includegraphics[width=\linewidth]{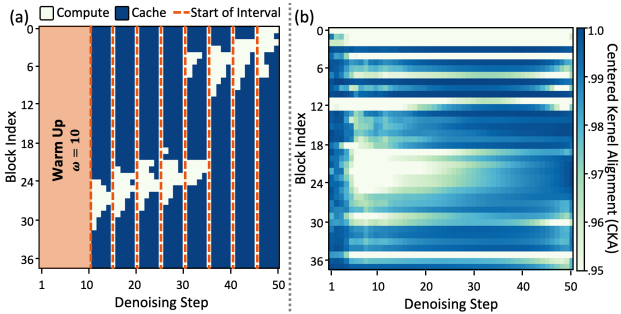}
\caption{\textbf{Ablation on contribution score.} Visualization of (a) CorGi's caching pattern and (b) the corresponding block-wise contribution analysis.}
\label{fig:ablation3}   
\vspace{-5pt}
\end{figure}

\subsection{Impact of Contribution Score}
\label{sec:ablation4}

Corgi uses a CKA-based contribution score to compute each block's contribution at initial step of each interval. Then, CorGi performs block-wise caching based on these scores. Fig.~\ref{fig:ablation3} visualizes (a) the caching pattern of CorGi and (b) the corresponding block-wise contribution analysis for the same prompt. As shown in Fig.~\ref{fig:ablation3}, CorGi effectively caches low-contribution blocks at each interval while preserving high-contribution blocks for computation. In addition, as the denoising step progresses, the important blocks gradually shift from deeper layers to shallower layers, and CorGi successfully tracks this shift when selecting blocks to cache.

\begin{figure}[!t]
\centering
\includegraphics[width=0.92\linewidth]{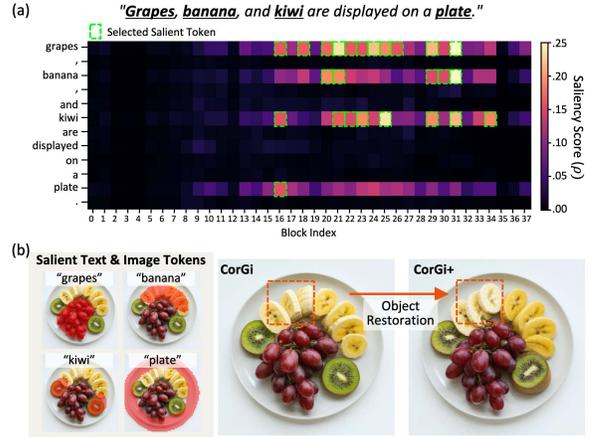}
\caption{\textbf{Ablation on salient token identification.} CorGi+ accurately identifies salient tokens in each block, enabling it to restore degraded objects.}
\label{fig:ablation4}   
\end{figure}

\subsection{Impact of Salient Token Identification}
\label{sec:ablation5}

After the $\omega$ warm-up steps, CorGi+ computes a saliency score $\rho$ for each token from the cross-attention maps and identifies per-block salient tokens based on these scores. It then performs partial updates on these salient tokens within cached blocks, thereby preserving fine object details in the generated images. Fig.~\ref{fig:ablation4} illustrates how CorGi+ identifies these salient tokens and restores object details that are degraded in CorGi. As shown in Fig.~\ref{fig:ablation4}(a), CorGi+ selects the top-$c$ salient tokens for each block based on their saliency scores. It then identifies the corresponding image tokens associated with these salient text tokens via $k$-means clustering, as shown in Fig.~\ref{fig:ablation4}(b). By performing partial update only on these block-specific salient tokens, CorGi+ successfully restores the \textit{banana} object details.

\section{Additional Qualitative Results}
\label{sec:appendix2}

\begin{figure*}[t]
\centering
\includegraphics[width=0.91\linewidth]{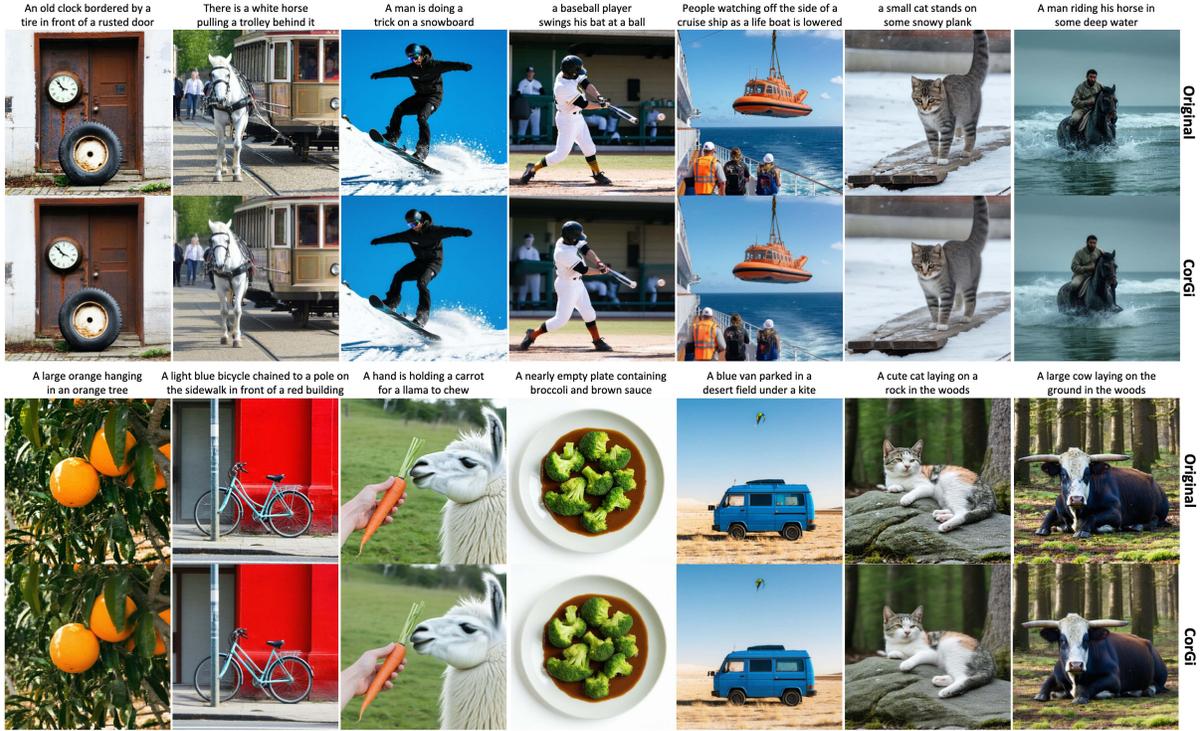}
\caption{\textbf{Qualitative results of CorGi on Stable Diffusion 3.5 Large.} CorGi achieves 2.09$\times$ speedup while maintaining strong consistency with the original images.}
\label{fig:qualitative1}   
\end{figure*}

As indicated in Fig.~\ref{fig:qualitative1}, ~\ref{fig:qualitative2}, ~\ref{fig:qualitative3}, ~\ref{fig:qualitative4}, ~\ref{fig:qualitative5}, and ~\ref{fig:qualitative6}, we provide additional qualitative results for both CorGi and CorGi+ across all evaluated DiT models -- Stable Diffusion 3.5 Large, FLUX.1-dev, and PixArt-$\Sigma$. These examples further illustrate that our methods preserve the visual and semantic fidelity to the outputs of the original model, while achieving substantial inference speedups. All prompts in theses qualitative examples are taken from the MS-COCO 2014 validation set.

\begin{table}[t!]
\centering
\resizebox{\columnwidth}{!}{
\begin{tabular}{l | c | cccc}
\toprule
\textbf{Method} & Speedup & FID ↓ & Pick ↑ & IR ↑ & LPIPS ↓ \\
\midrule
FLUX.1-dev, 30 steps & \textcolor{customblue}{--} 
& 30.53 & 23.13 & 1.09 & -- \\
\midrule
DeepCache ($N$=2) & \textcolor{customblue}{\textbf{1.26$\times$}} 
& 29.32 & 23.12 & 1.09 & 0.25 \\
BlockDance ($N$=3)& \textcolor{customblue}{\textbf{1.50$\times$}} 
& 29.22 & 23.14 & 1.10 & 0.05 \\
RAS (50\% ratio)& \textcolor{customblue}{\textbf{1.98$\times$}}
& 30.92 & 22.78 & 1.01 & 0.14 \\
\midrule
\rowcolor[gray]{0.95}
CorGi ($D$=5, $\gamma$=49, $\delta$=2) & \textcolor{customblue}{\textbf{2.00$\times$}} 
& 31.33 & 22.99 & 1.06 & 0.09 \\
\rowcolor[gray]{0.95}
CorGi ($D$=6, $\gamma$=47, $\delta$=2) & \textcolor{customblue}{\textbf{2.08$\times$}} 
& 30.83 & 22.89 & 1.06 & 0.10 \\
\midrule
\rowcolor[gray]{0.9}
CorGi+ ($D$=5, $\gamma$=49, $\delta$=2) & \textcolor{customblue}{\textbf{1.91$\times$}} & 30.40 & 22.97 & 1.05 & 0.09 \\
\rowcolor[gray]{0.9}
CorGi+ ($D$=6, $\gamma$=47, $\delta$=2) & \textcolor{customblue}{\textbf{1.98$\times$}} & 30.24 & 22.84 & 1.01 & 0.12 \\
\midrule
\rowcolor[gray]{0.85}
CorGi+ ($D$=5, $\gamma$=36, $\delta$=2) & \textcolor{customblue}{\textbf{1.53$\times$}} & 29.77 & 23.04 & 1.08 & 0.08 \\
\rowcolor[gray]{0.85}
CorGi+ ($D$=6, $\gamma$=33, $\delta$=2) & \textcolor{customblue}{\textbf{1.54$\times$}} & 28.85 & 22.95 & 1.08 & 0.10 \\
\bottomrule
\end{tabular}
}
\caption{\textbf{Text-to-image generation comparison on FLUX.1-dev.}}
\label{tab:flux_additional}
\end{table}

\section{Additional Experiments}
\label{sec:appendix3}

We further evaluate CorGi+ on FLUX.1-dev using 5K prompts from the MS-COCO 2014 validation set. To analyze the speed-quality trade-off, we evaluate CorGi+ with a lower caching ratio. As shown in Table~\ref{tab:flux_additional}, CorGi and CorGi+ outperform all baselines in FID at similar speedup levels, while maintaining comparable performance on the other metrics. For example, CorGi+ (D=6, $\gamma$=33, $\delta$=2) achieves lower FID than both DeepCache and BlockDance. Similarly, CorGi (D=6, $\gamma$=47, $\delta$=2) outperforms RAS across all evaluation metrics. In all these configurations, CorGi and CorGi+ still remain faster than the corresponding baselines. These results indicate that CorGi and CorGi+ achieve a better speed-quality trade-off, as their finer block-level caching removes redundant computation more effectively than existing baselines. 

\begin{figure*}[t]
\centering
\includegraphics[width=0.91\linewidth]{Figure/Figure_qualitative2.pdf}
\caption{\textbf{Qualitative results of CorGi on FLUX.1-dev.} CorGi achieves 2.08$\times$ speedup while maintaining strong consistency with the original images.}
\label{fig:qualitative2}   
\end{figure*}

\begin{figure*}[t]
\centering
\includegraphics[width=0.91\linewidth]{Figure/Figure_qualitative3.pdf}
\caption{\textbf{Qualitative results of CorGi on PixArt-$\Sigma$.} CorGi achieves 2.00$\times$ speedup while maintaining strong consistency with the original images.}
\label{fig:qualitative3}   
\end{figure*}

\begin{figure*}[t]
\centering
\includegraphics[width=0.91\linewidth]{Figure/Figure_qualitative4.pdf}
\caption{\textbf{Qualitative results of CorGi+ on Stable Diffusion 3.5 Large.} CorGi+ achieves 2.02$\times$ speedup while maintaining strong consistency with the original images.}
\label{fig:qualitative4}   
\end{figure*}

\begin{figure*}[t]
\centering
\includegraphics[width=0.91\linewidth]{Figure/Figure_qualitative5.pdf}
\caption{\textbf{Qualitative results of CorGi+ on FLUX.1-dev.} CorGi+ achieves 1.91$\times$ speedup while maintaining strong consistency with the original images.}
\label{fig:qualitative5}   
\end{figure*}

\begin{figure*}[t]
\centering
\includegraphics[width=0.91\linewidth]{Figure/Figure_qualitative6.pdf}
\caption{\textbf{Qualitative results of CorGi+ on PixArt-$\Sigma$.} CorGi+ achieves 1.97$\times$ speedup while maintaining strong consistency with the original images.}
\label{fig:qualitative6}   
\end{figure*}

\end{document}